\documentstyle[aclap]{article}

\title{\vspace{-0.5in}Conciseness through Aggregation in Text Generation}
\author{James Shaw \\
Dept.~of Computer Science\\
Columbia University \\
New York, NY 10027, USA \\
{\tt shaw@cs.columbia.edu}}

\begin{document}

\maketitle

\begin{abstract}

Aggregating different pieces of similar information is
necessary to generate concise and easy to understand reports in
technical domains.  This paper presents a general algorithm that
combines similar messages in order to generate one or more coherent
sentences for them.  The process is not as trivial as might be
expected.  Problems encountered are briefly described.
\end{abstract}

\vspace{-0.1in}
\section{Motivation}
\vspace{-0.07in}

Aggregation is any syntactic process that allows the expression of
concise and tightly constructed text such as coordination or
subordination.  By using the parallelism of syntactic structure to
express similar information, writers can convey the same amount of
information in a shorter space.  Coordination has been the object of
considerable research (for an overview, see \cite{van Oirsouw87}).  In
contrast to linguistic approaches, which are generally analytic, the
treatment of coordination in this paper is from a synthetic point of
view --- text generation.  It raises issues such as
deciding when and how to coordinate.  An algorithm for generating
coordinated sentences is implemented in PLANDoc \cite{Kukich et
al. 93,McKeown et al. 94}, an automated documentation system.

PLANDoc generates natural language reports based on the interaction
between telephone planning engineers and LEIS-PLAN\footnote{LEIS is a
registered trademark of Bell Communications Research, Piscataway,
NJ.}, a knowledge based system.  Input to PLANDoc is a series of
messages, or semantic functional descriptions (FD, Fig. 1).  Each FD
is an atomic decision about telephone equipment installation chosen by
a planning engineer.  The domain of discourse is currently limited to
31 message types, but user interactions include many variations and
combinations of these messages.  Instead of generating four separate
messages as in Fig. 2, PLANDoc combines them and generates the
following two sentences: ``{\em This refinement activated DLC for CSAs
3122 and 3130 in the first quarter of 1994 and ALL-DLC for CSA 3134 in
1994 Q3.  It also activated DSS-DLC for CSA 3208 in 1994 Q3.}''

\begin{figure}[b]
\vspace{-0.2in}
\small
\rule{3.1in}{.01in}
\vspace{-0.2in}
\begin{verbatim}
((cat message)
 (admin ((PLANDoc-message-name RDA)
         (runid r-reg1)))
 (class refinement)
 (action activation)
 (equipment-type  all-dlc)
 (csa-site 3134)
 (date ((year 1994) (quarter 3))))
\end{verbatim}
\vspace{-0.1in}
\caption{Output of the Message Generator}
\rule{3.1in}{.01in}
\vspace{-0.3in}
\end{figure}

\begin{figure*}
\small
\rule{\textwidth}{.01in}
\begin{verbatim}
This refinement activated ALL-DLC for CSA 3134 in 1994 Q3.  (E1 S3 D2)
This refinement activated DLC for CSA 3130 in 1994 Q1.      (E2 S2 D1)
This refinement activated DSS-DLC for CSA 3208 in 1994 Q3.  (E3 S4 D2)
This refinement activated DLC for CSA 3122 in 1994 Q1.      (E2 S1 D1)
\end{verbatim}
\vspace{-0.1in}
\scriptsize
\begin{verbatim}
Equipment: E1= ALL-DLC, E2= DLC, E3= DSS-DLC
         Site: S1= CSA 3122, S2= CSA 3130, S3= CSA 3134, S4= CSA 3208
        Date: D1= l994 Q1, D2= 1994 Q3
\end{verbatim}
\vspace{-0.2in}
\caption{Unaggregated Text Output}
\rule{\textwidth}{.01in}
\end{figure*}

\section{System Architecture}
\vspace{-0.07in}

Fig. 3 is an overview of PLANDoc's architecture.  Input to the message
generator comes from LEIS-PLAN tracking files which record user's
actions during a planning session.  The ontologizer adds hierarchical
structure to messages to facilitate further processing.  The
content planner organizes the overall narrative and determines the
linear order of the messages.  This includes combining atomic messages
into aggregated messages, choosing cue words, and determining
paraphrases that maintain focus and ensure coherence.  Finally the
FUF/SURGE package \cite{Elhadad91,Robin-PhD} lexicalizes the messages
and maps case roles into syntactic roles, builds the constituent
structure of the sentence, ensures agreement, and generates the
surface sentences.

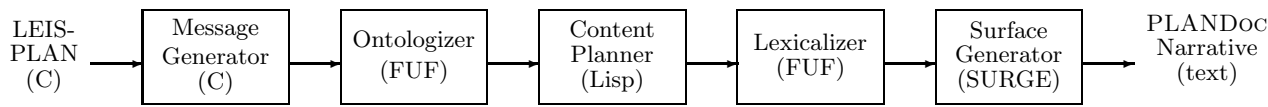
\begin{figure*}
\rule{\textwidth}{.01in}
\small
\begin{picture}(500,40)
\put(-20,0){\makebox(55,35){\shortstack{LEIS-\\PLAN\\(C)}}}
\put(45,0){\framebox(55,35){\shortstack{Message\\Generator\\(C)}}}
\put(120,0){\framebox(55,35){\shortstack{Ontologizer\\(FUF)}}}
\put(195,0){\framebox(55,35){\shortstack{Content\\Planner\\(Lisp)}}}
\put(270,0){\framebox(55,35){\shortstack{Lexicalizer\\(FUF)}}}
\put(345,0){\framebox(55,35){\shortstack{Surface\\Generator\\(SURGE)}}}
\put(420,0){\makebox(55,40){\shortstack{{\sc PLANDoc}\\Narrative\\(text)}}}

\put(25,15){\vector(1,0){20}}
\put(100,15){\vector(1,0){20}}
\put(175,15){\vector(1,0){20}}
\put(250,15){\vector(1,0){20}}
\put(325,15){\vector(1,0){20}}
\put(400,15){\vector(1,0){20}}
\end{picture}

\normalsize
\vspace{-0.1in}
\caption{PLANDoc System Architecture}
\rule{\textwidth}{.01in}
\vspace{-0.3in}
\end{figure*}

\normalsize

\section{Combining Strategy}
\vspace{-0.07in}

Because PLANDoc can produce many paraphrases for a single message,
aggregation during the syntactic phase of generation would be
difficult; semantically similar messages would already have different
surface forms.  As a result, aggregation in PLANDoc is carried out at
the content planning level using semantic FDs.  Three main criteria
were used to design the combining strategy:

\begin{enumerate}
\item {\bf domain independence}: the algorithm should be applicable in
other domains.
\item {\bf generating the most concise text}: it should avoid
repetition of phrases to generate shortest text.
\item {\bf avoidance of overly-complex sentences}: it should not
generate sentences that are too complex or ambiguous for readers.
\end{enumerate}
\small
\normalsize

\noindent
The first aggregation step is to identify semantically related
messages.  This is done by grouping messages with the same action
attribute.  Then the system attempts to generate concise and
unambiguous text for each action group separately.  This reduces the
problem size from tens of messages into much smaller
sizes.  Though this heuristic disallows the combination of messages
with different actions, the messages in each action group already
contain enough information to produce quite complex sentences.

The system combines the maximum number of related messages to meet the
second design criterion--generating the most concise text.  But such
combination is blocked when a sentence becomes too complex.  A
bottom-up 4-step algorithm was developed:

\begin{enumerate}
\small
\normalsize
\item {\bf Sorting}: putting similar messages right next to each other.
\item {\bf Merging Same Attribute}: combining adjacent messages that
only have one distinct attribute.
\item {\bf Identity Deletion}: deletion of identical components across
messages.
\item {\bf Sentence Breaking}: determining sentence breaks.
\end{enumerate}
\small
\normalsize
\vspace{-0.1in}
\noindent

\subsection{Step 1: Sorting}

The system first ranks the attributes to determine which
are most similar across messages with the same action.  For each
potential distinct attribute, the system calculates its rank using the
formula $m - d$, where $m$ is the number of messages and $d$ is the
number of distinct attributes for that particular attribute.
The rank is an indicator of how similar an attribute
is across the messages.  Combining messages according to the highest
ranking attribute ensures that minimum text will be generated for
these messages.
Based on the ranking, the system reorders the
messages by sorting, which puts the messages that have the same
attribute right next to each other.
In Fig. 2, {\em equipment} has rank 1 because it has 3 distinct
equipment values -- ALL-DLC, DLC, and DSS-DLC; {\em date} has rank 2
because it has two distinct date values -- 1994 Q1 and 1994 Q3; {\em
site} has rank 0.  Attribute {\em class} and {\em action} (Fig. 1) are
ignored because they are always the same at this stage.  When two
attributes have the same rank, the system breaks the tie based on a
priority hierarchy determined by the domain experts.  Because the
final sorting operation dominates the order of the resulting messages,
PLANDoc sorts the message list from the lowest rank attribute to the
highest.  In this case, the ordering for sorting is {\em site}, {\em
equipment}, and then {\em date}.
The resulting message list after sorting each
attribute is shown in Fig. 4.

\begin{figure}
\small
\rule{3.1in}{.01in}
\begin{verbatim}
  (E2 S1 D1)     (E1 S3 D2)     (E2 S1 D1)
  (E2 S2 D1)     (E2 S1 D1)     (E2 S2 D1)
  (E1 S3 D2) --> (E2 S2 D1) --> (E1 S3 D2)
  (E3 S4 D2)     (E3 S4 D2)     (E3 S4 D2)
    by Site     by Equipment      by Date
\end{verbatim}
\vspace{-0.1in}
\caption{Step 1.  Sorting}
\rule{3.1in}{.01in}
\vspace{-0.4in}
\end{figure}

\subsection{Step 2: Merging Same Attribute}

The list of sorted messages is traversed.  Whenever there is only one
distinct attribute between two adjacent messages, they are merged into
one message with a conjoined attribute, which is a list of the
distinct attributes from both messages.  What about messages with two
or more distinct attributes?  Merging two messages with two or more
distinct attributes will result in a syntactically valid sentence but
with an undesirable meaning: ``{\em *This refinement activated ALL-DLC
and DSS-DLC for CSAs 3122 and 3130 in the third quarter of 1993.}''

By tracking which attribute is compound, a third message can be merged
into the aggregate message if it also has the same distinct
attribute.  Continue from Step 1, (E2 S1 D1) and (E2
S2 D1) are merged because they have only one distinct attribute, {\em site}.
A new FD, (E2 (S1 S2) D1), is assembled to replace those two
messages.  Note that although (E1 S3 D2) and (E3
S4 D2) have the date in common, they are not combined because they have
more than one distinct attribute, {\em site} and {\em equipment}.

Step 2 is applied to the message list recursively to generate possible
crossing conjunction, as in the following output which merges {\em
four} messages: ``{\em This refinement activated ALL-DLC and DSS-DLC
for CSAs 3122 and 3130 in the third quarter of 1993.}''  Though on the
outset this phenomenon seems unlikely, it does happen in our domain.

\subsection{Step 3: Identity Deletion}
After merging at step 2, the message list left in an action group
either has only one message, or it has more than one message with at
least two distinct attributes between them.  Instead of generating two
separate sentences for (E2 (S1 S2) D1) and (E1 S3 D2), the system
realizes that both the subject and verb are the same, thus it uses
deletion on identity to generate ``{\em This refinement activated DLC
for CSAs 3122 and 3130 in 1994 Q1 and [this refinement activated]
ALL-DLC for CSA 3134 in 1994 Q3.}''  For identical attributes across
two messages (as shown in the bracketed phrase), a ``deletion''
feature is inserted into the semantic FD, so that SURGE will suppress
the output.

\subsection{Step 4: Sentence Break}

Applying deletion on identity blindly to the whole message list might make the
generated text incomprehensible because readers might have to recover
too much implicit information from the sentence.  As a result, the
combining algorithm must have a way to determine when to break the
messages into separate sentences that are easy to understand and
unambiguous.

How much information to pack into a sentence does not depend on
grammaticality, but on coherence, comprehensibility, and aesthetics
which are hard to formalize.
PLANDoc uses a heuristic that always joins the first and
second messages, and continues to do so for third and more if the
distinct attributes between the messages are the same.
This heuristics results in parallel syntactic structure and the
underlying semantics can be easily recovered.  Once the distinct
attributes are different from the combined messages, the system starts
a new sentence.  Using the same example, (E2 (S1 S2) D1) and (E1 S3
D2) have three distinct attributes.  They are combined because they
are the first two messages.  Comparing the third message (E3 S4 D2) to
(E1 S3 D2), they have different {\em equipment} and {\em site}, but
not {\em date}, so a sentence break will take place between them.
Aggregating all three messages together will results in questionable
output.  Because of the parallel structure created between the first 2
messages, readers are expecting a different {\em date} when
reading the third clause.  The second occurrence of ``1994 Q3'' in
the same sentence does not agree with readers' expectation thus
potentially confusing.

\section{Future Directions}
\vspace{-0.07in}

In this paper, I have described a general algorithm which not only
reduces the amount of the text produced, but also increases the
fluency of the text.  While other systems do generate conjunctions,
they deal with restricted cases such as conjunction of subjects and
predicates\cite{Dalianis}.  There are other interesting problems in
aggregations.  Generating marker words to indicate relationships in
conjoined structures, such as ``respectively'', is another short term
goal.  Extending the current aggregation algorithm to be more general
is currently being investigated, such as combining related messages
with different actions.

\section{Acknowledgements}
\vspace{-.07in}

The author thanks Prof. Kathleen McKeown, and Dr. Karen Kukich at
Bellcore for their advice and support.  This research was conducted
while supported by Bellcore project \#CU01403301A1, and under the
auspices of the Columbia University CAT in High Performance Computing
and Communications in Healthcare, a New York State Center for Advanced
Technology supported by the New York State Science and Technology
Foundation.

\end{document}